# Learning to Order: Task Sequencing as In-Context Optimization

Jan Kobiolka[1] (✉), Christian Frey[1], Arlind Kadra[1], Gresa Shala[2], and Josif Grabocka[1]

[1] Department of Computer Science and Artificial Intelligence, University of Technology Nuremberg, Germany `jan.kobiolka@utn.de, christian.frey@utn.de, arlind.kadra@utn.de, josif.grabocka@utn.de`
[2] Department of Computer Science, Albert Ludwig University of Freiburg, Germany `shalag@cs.uni-freiburg.de`

**Abstract.** Task sequencing (TS) is one of the core open problems in Deep Learning, arising in a plethora of real-world domains, from robotic assembly lines to autonomous driving. Unfortunately, prior work has not convincingly demonstrated the generalization ability of meta-learned TS methods to solve new TS problems, given few initial demonstrations. In this paper, we demonstrate that deep neural networks can meta-learn over an infinite prior of synthetically generated TS problems and achieve a few-shot generalization. We meta-learn a transformer-based architecture over datasets of sequencing trajectories generated from a prior distribution that samples sequencing problems as paths in directed graphs. In a large-scale experiment, we provide ample empirical evidence that our meta-learned models discover optimal task sequences significantly quicker than non-meta-learned baselines.

## 1 Introduction

Task sequencing (TS) is the process of arranging tasks in a specific order to enable gradual and effective knowledge acquisition. This concept arises in domains such as human skill acquisition [5], game design and progression planning [12], and humancomputer interaction (HCI) [30], where task order can strongly influence learning efficiency and performance outcomes. More recently, task sequencing has been deployed in language models [22,16], reinforcement learning [24,23], and autonomous driving [32]. In optimization theory, ordered task presentation has been shown to accelerate convergence in non-convex problems and yield better local optima [7]. However, a persistent limitation of prior work is the reliance on heuristic rules [2,3], domain-specific knowledge [18], or handcrafted curricula [13], which restricts applicability across domains. To date, no prior work has demonstrated that deep neural networks can be meta-learned over a large prior of TS problems and generalize to new TS problems with only a few example sequences. In other words, no attempt exists for foundation models in the realm of task sequencing.

In this paper, we empirically demonstrate that deep neural networks can successfully meta-learn from a large prior of diverse TS problems. Considering the lack of a public large-scale repository for real-world task sequencing problems, we designed a novel prior distribution over TS problems, from which we can sample infinitely many synthetic sequencing problems. Afterwards, we formulate a meta-learning objective for in-context task sequencing that learns to predict an optimal task sequence (target) given a few non-optimal sequences (context). Each meta-learning batch represents sequences from a newly sampled TS problem, where the latter consists of task paths in a task graph with a unique adjacency matrix for each sequencing problem. Furthermore, we propose a utility function to label the optimality of each sequence, defined as the sequence's similarity to one of the optimal solutions of the sampled TS problem. As an overarching goal, we relax the dependency on expert-designed curricula and instead learn to sequence tasks in a domain-agnostic manner using in-context learning.

To empirically demonstrate that our model generalizes in solving few-shot TS problems, we designed a large-scale empirical protocol consisting of hundreds of thousands of sampled sequencing problems. As the most important research question, we show that our approach outperforms non-meta-learned baselines in searching for optimal sequences on a hold-out set of hundreds of testing TS problems. In addition, our model is capable of approximating optimal sequences across various sequence lengths, outperforming rule-based, RL and large language model baselines. Overall, we make the following contributions:

– We propose a meta-learned, in-context model that learns to sequence tasks by approximating optimal orderings based on a few context sequences. Our method is domain-agnostic and generalizes to arbitrary task priors.
– We introduce a universal graph-based task prior, generated via stochastic graph expansion, that encodes diverse topologies and learning trajectories.
– We empirically demonstrate that our model successfully predicts optimal sequences automatically in a few-shot learning scenario, outperforming rule-based, RL and large language model baselines.

## 2   Prior-Fitted Task Sequencing Networks (PFTSN)

**Problem Definition.** Let $\mathcal{T} := \{t_1, \ldots, t_N\}$ denote a set of $N$ tasks, and let a sequence of $L$ tasks be an ordered list $\tau \in \mathcal{T}^L$. The goal of TS is to identify an optimal sequence $\tau^*$ that maximizes the performance on a target task. We denote a TS problem as $\mathcal{P} := \langle S^*, U \rangle$, where $S^* \subseteq \mathcal{T}^L$ is a set of optimal sequences known for the problem instance, and the utility function $U(\tau) : \mathcal{T}^L \to \mathbb{R}_+^L$ measures performance after having trained on each of the tasks of a sequence $\tau$. Concretely, $U_k(\tau)$ denotes the utility after executing the first $k$ tasks in $\tau$, i.e., for the prefix $\tau_{1:k}$. Since $U(\tau)$ is vector-valued, we define a scalar utility via $\bar{U}(\tau) := \sum_{k=1}^{L} U_k(\tau)$. The optimal set of sequences $S^*$ satisfies

$\forall \tau^* \in S^*,\ \forall \tau \in \mathcal{T}^L \setminus S^* : \bar{U}(\tau) < \bar{U}(\tau^*)$, i.e., every sequence in $S^*$ is optimal under $\bar{U}$.

On the other hand, an oracle task sequence solver is a probability distribution $\tau^* \sim p(\tau^* \mid S, \theta)$ that can be queried to sample an optimal sequence $\tau^* \in S^*$ given a few observations $S$ of non-optimal sequences and their corresponding utility values. We denote by $U(S) := \{U(\tau) \mid \tau \in S\}$ the set of observed utility vectors, and use $(S, U(S))$ as contextual information. Assume that we have access to a prior over all sequencing problems in the universe, denoted as $\mathcal{P} \sim p(\mathcal{P})$, and we could sample infinitely many task sequencing problems as $\langle S^*, U \rangle \sim p(\mathcal{P})$. In that case, we could fit parametric neural networks as task sequencing policies $\tau^* \sim \pi(\tau^* \mid S, U(S); \theta)$ via negative likelihood estimation to jointly solve all TS problems in our prior distribution as:

$$\theta^* \in \arg\min_\theta\ \mathbb{E}_{\langle S^*, U \rangle \sim p(\mathcal{P})}\ \mathbb{E}_{S \subseteq \{\tau \mid \bar{U}(\tau) < \bar{U}(\tau^*),\ \tau^* \in S^*\}}\ \min_{\tau^* \in S^*}\ -\log \pi(\tau^* \mid S, U(S); \theta). \quad (1)$$

We optimize the parameters $\theta$ in expectation over all the problems from our prior distribution $p(\mathcal{P})$. Section 3 describes the prior distribution, how we can sample task sequencing problems and introduces the utility function. Once we sample a problem $\langle S^*, U \rangle$, we can train our model using supervised learning by predicting the closest optimal trajectory $\tau^*$ using as input a set of a few non-optimal trajectories $S$ together with their utilities $U(S)$. The reason for using the $\min_{\tau^* \in S^*}$ is because we have non-unique optimal trajectories in $S^*$; therefore, we choose the closest optimal sequence to our model's prediction (i.e., smallest loss). The objective function of Equation 1 maximizes the likelihood of optimal sequences over all the task sequencing problems from $p(\mathcal{P})$.

**Algorithm:** Meta-Learning PFTSN
1: **input:** Prior distribution over TS problems $p(S^*, U)$
2: **for** update steps $t = 1, \ldots$ **do**
3:     Sample TS problem: $\langle S^*, U \rangle \sim p(S^*, U)$
4:     Sample random sequences $S_{\text{rand}} \sim \text{Uniform}\left(\mathcal{T}^L\right)^{C_{\text{rand}}}$
5:     Sample mutated sequences $S_{\text{mut}} \sim \text{Mutate}(S^*)$
6:     Batch of non-optimal sequences: $S \leftarrow S_{\text{rand}} \cup S_{\text{mut}}$
7:     Update: $\theta \leftarrow \theta - \eta \frac{\partial \min_{\tau^* \in S^*} [-\log \pi(\tau^* \mid S, U(S); \theta)]}{\partial \theta}$
8: **end for**
9: **return:** Model parameters $\theta$

**Training procedure.** We generate training batches using the prior over TS problems. First, we sample a problem $\langle S^*, U \rangle \sim p(S^*, U)$. Afterwards, we generate a batch of context sequences $S$ using two approaches: random sampling and mutating optimal sequences. We sample $C_{\text{rand}} \in \mathbb{N}$ random sequences as $S_{\text{rand}} \sim \text{Uniform}\left(\mathcal{T}^L\right)^{C_{\text{rand}}}$. Afterwards, we sample $C_{\text{mut}} \in \mathbb{N}$ mutated sequences by sampling optimal sequence from $S^*$ and permuting multiple random pairs of tasks on the optimal sequence $S_{\text{mut}} \sim \text{Mutate}(S^*)$ (more information in Appendix A). We concatenate the batches of random and mutated sequences as $S \leftarrow S_{\text{rand}} \cup S_{\text{mut}}$. Our method meta-learns the TS network parameters to estimate the closest optimal sequence $\tau^*$ given the sequences $S$ and their utility $U(S)$ values as network inputs, in expectation over TS problems $\langle S^*, U \rangle$ that we sample from the prior. We dub our method PFTSN - Prior-fitted Task Sequencing Network.

**Model architecture.** We concatenate $S$, $U(S)$, and a normalized positional information vector (pos) into $X \in \mathbb{R}^{B \times C \times L \times 3}$, where $B$ is the batch size, $C = C_{\text{rand}} + C_{\text{mut}}$, $L$ the sequence length, and the last dimension stacks $(S, U(S), \text{pos})$ per position. We apply a linear transformation $X' = XW_{in}$, where $W_{in} \in \mathbb{R}^{d \times d_{\text{emb}}}$ is a learnable weight matrix. In our architecture, we deploy $K$ alternating sequence and task attention [29] (starting and ending with a sequence attention block) for a context tensor $X'$ as follows:

$$X_{\text{seq}} = \text{RMSNorm}(X' + \text{MultiHeadAttention}(\text{RMSNorm}(X'))) \quad (2)$$

$$X_{\text{task}} = \text{RMSNorm}(X_{\text{seq}}^T + \text{MultiHeadAttention}(\text{RMSNorm}(X_{\text{seq}}^T)))^T \quad (3)$$

where $\text{MultiHeadAttention}(Z) = \text{Concat}(\text{head}_1, \ldots, \text{head}_H)W_o$ for an input matrix $Z \in \mathbb{R}^{B \times C \times L \times d_{emb}}$. Each head applies a standard self-attention $\text{head}_i = \text{Softmax}((Q_i K_i)/\sqrt{d_{\text{head}}})V_i$, with $Q_i = ZW_i^Q, K_i = ZW_i^K, V_i = ZW_i^V$ and $d_{\text{head}} = d_{\text{emb}}/H$. As feedforward networks, we apply $\text{FFN}(Z) = W_2(\text{SiLU}(W_1 Z) \odot W_3 Z)$ at the end of each attention block with $W_1, W_2, W_3 \in \mathbb{R}^{d_{emb} \times d_{emb}}$ being learnable weights, and $\odot$ denoting an element-wise multiplication [27]. After the operations, we obtain a vector of $X' \in \mathbb{R}^{B \times C \times L \times d_{emb}}$. After applying RMS normalization [33], we apply mean pooling over $C$ yielding $X_{att} \in \mathbb{R}^{B \times L \times d_{emb}}$.

In order to auto-regressively generate a new sequence given context $X$, we sample one optimal trajectory $\tau^*$ per batch size $B$. This trajectory is then prefixed with a <BOS> token (omitting its final token). This sequence serves as our current trajectory for which we require the next task. After stacking an <unknown> token for the similarity vector and appending a positional information vector, we obtain a tensor of shape $T \in \mathbb{R}^{B \times 1 \times L \times 3}$. We project it linearly via $T' = TW_{in}$. Next, we apply $K$ task attention layers with causal masking to process the information contained within the current sequence, yielding $T_{att}$ to which we also apply RMS normalization. The predicted task is obtained via $\hat{y} = \text{FFN}([T_{att}||X_{att}])$, where $||$ denotes vector concatenation and FFN is a feed-forward neural network mapping from $\mathbb{R}^{2 \cdot d_{\text{emb}}}$ to the task output space $\mathbb{R}^{|\mathcal{T}|}$. We generate a new task from the obtained logits by sampling with the softmax function. Figure 1 illustrates the general workflow of our architecture.

## 3   A Prior Distribution for Task Sequencing Problems

We previously explained how we train prior-fitted task sequencing networks given a prior over TS problems $p(S^*, U)$. This section details our sampling approach based on task graphs. Section 3.1 describes optimal sequence $S^*$ generation, while Section 3.3 defines the utility function $U(\tau)$.

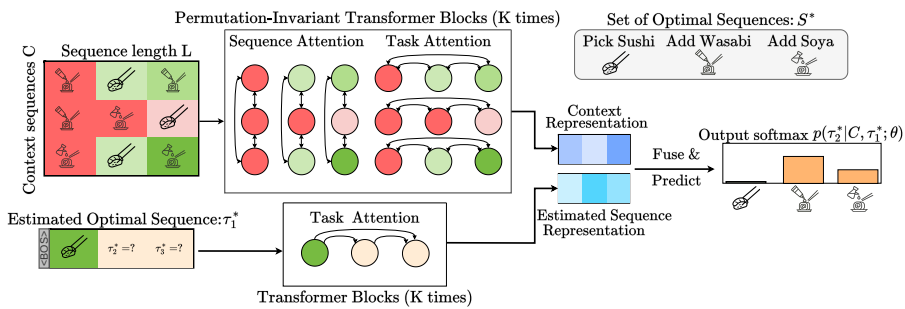

Fig. 1: Workflow of our PFTSN architecture.

### 3.1  Generating Optimal Sequences $S^*$

A curriculum yields an ordering of tasks that is provided to the learner. In our work, we define our prior as graph traversals that define a set of optimal task sequences.

**Definition 1 (Task Graph).** *Let $\mathcal{T} \coloneqq \{t_1, \ldots, t_N\}$ denote a set of $|\mathcal{T}|=N$ tasks. We define a directed acyclic graph (DAG) $\mathcal{G}=(\mathcal{V}, \mathcal{A})$, where each node $v \in \mathcal{V}$ corresponds to a task via a mapping $\phi : \mathcal{V} \to \mathcal{T}$. The edges $(v_i, v_j) \in \mathcal{A}$ denote transitions or dependencies between tasks in the traversals. Let $\delta^-(v_i) = \{v_h | \forall v_h \in \mathcal{V} : (v_h, v_i) \in \mathcal{A}\}$ and $\delta^+(v_i) = \{v_h | \forall v_h \in \mathcal{V} : (v_i, v_h) \in \mathcal{A}\}$ denote the in-arcs and out-arcs for a node $v_i$, respectively. We define the set of nodes $R(\mathcal{G}) \coloneqq \{v_i | \forall v_i \in \mathcal{V} : \delta^-(v_i) = \emptyset\}$ as the root nodes of $\mathcal{G}$, and the leaf nodes as $D(\mathcal{G}) \coloneqq \{v_i | \forall v_i \in \mathcal{V} : \delta^+(v_i) = \emptyset\}$.*

We allow $\phi(\cdot)$ to assign the same task $t_i \in \mathcal{T}$ to multiple nodes in $\mathcal{V}$, enabling repeated presentation of tasks within a task sequence. Given a task-graph $\mathcal{G}$, we define the optimal task sequences $S^*$ as all paths from the designated start nodes $r_i \in R$ nodes to leaf nodes $d_i \in D$, hence, reflecting a set of valid task sequences of the task set $T$. We define:

**Definition 2 (Task Path).** *Give a task sequencing problem as $\mathcal{P} \coloneqq \langle S^*, U \rangle$, we define each optimal sequence $\tau \in S^*$ of length $L$ as an ordered lists of tasks $\tau = (\tau_1, \ldots, \tau_L)$ such that there exists a corresponding path $(v_1, \ldots, v_L)$ in its respective task-graph $\mathcal{G}$. Thus, the set of optimal sequences satisfies $S^* \coloneqq \{\tau = (\tau_1, \ldots, \tau_L) | \exists (v_1, \ldots, v_L) \in \mathcal{V}^L : ((v_i, v_{i+1}) \in \mathcal{A}) \wedge (\tau_i = \phi(v_i)) \wedge (\tau_{i+1} = \phi(v_{i+1}))\}$. Analogously, we denote by $S \coloneqq \{\tau \in \mathcal{T}^L | \tau \text{ is not a path in } \mathcal{G}, \tau \notin S^*\}$ the set of sequences of length $L$ which are non-optimal w.r.t. $\mathcal{G}$.*

Intuitively, Definition 2 ensures that a valid path sequence starts at any root node $r_i \in R$ of $\mathcal{G}$ and follows directed edges of the tree to a leaf node $d_i \in D$. A mapping function $\phi(\cdot)$ ensures that each node in the path is associated with a corresponding task. A poof that a DAG is universal for any finite sequencing problem is discussed in Section 3.2.

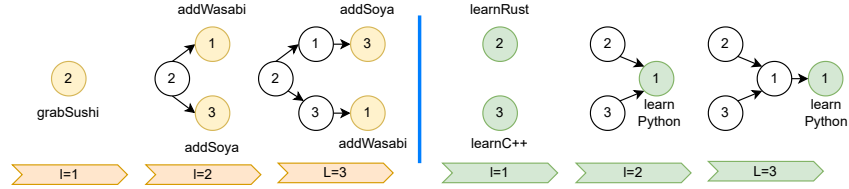

Fig. 2: Examples of generating optimal sequences $S^{*(orange)}$ (left) and $S^{*(green)}$ (right)

Next, we demonstrate how $S^*$ are constructed for a single sequencing problem $\mathcal{P} = \langle S^*, U \rangle$. Given a finite set of tasks $\mathcal{T} := \{t_1, \ldots, t_N\}$, we follow an iterative graph expansion procedure to generate task sequences $S^*$ denoting the optimal set for a task sequencing problem. We define $L$ as the maximal sequence length, and $k_{max} \in \mathbb{N}$ as the maximum number of nodes to add per step. For each step $0 < \ell < L$, we expand the current task graph $G_\ell$ at step $\ell$ by choosing one of the following graph expansion rules:

- *(Atomic Op.)* A single task $t_i \in \mathcal{T}$ is sampled uniformly at random. For the leaf nodes $D(\mathcal{G}_\ell)$, we create a new node $v'$ and define $\mathcal{V}_{\ell+1} := \mathcal{V}_\ell \cup \{v'\}$, $\mathcal{A}_{\ell+1} := \mathcal{A}_\ell \cup \{(v_i, v') | \forall v_i \in D(\mathcal{G}_\ell)\}$, with $\phi(v') = t_i$.
- *(Or Op.)* The or operation introduces parallel branches in $\mathcal{G}$. First, we sample $k = \min(k_{max}, L - |\tau|)$ distinct tasks $\mathcal{T}_k \subseteq \mathcal{T}$ uniformly at random. For each leaf node $v \in D(\mathcal{G}_\ell)$, we create a node $v_i$ for each sample $t_i \in \mathcal{T}_k$, i.e., $\mathcal{V}_{\ell+1} := \mathcal{V}_\ell \cup \{v_i | \forall i \in [1..k]\}$ with their links $\mathcal{A}_{\ell+1} := \mathcal{A}_\ell \cup \{(v, v_i) | \forall i \in [1..k]\}$ that maps to their respective tasks $\phi(v_i) = t_i$.
- *(And Op.)* The and operation creates multiple branches including permutation in task subsequences. We select $\mathcal{T}_k \subseteq \mathcal{T}$ as described for the *or* operation. Next, we generate all permutations of the selected tasks denoted as $\text{Perm}(\mathcal{T}_k)$. For each $\sigma = (t_{i_1}, \ldots, t_{i_k}) \in \text{Perm}(\mathcal{T}_k)$, we construct a path of new nodes $v_{i_1} \to \ldots \to v_{i_k}$, where $\phi(v_{i_j}) = t_{i_j}$ and the set of nodes and edges are expanded accordingly.

Notably, we sample the operations with equal probability to promote unbiased diversity among the generated task graphs. The generation process terminates when all sequences in $S^*$ reach the predefined length $L$, in other words, the number of edges from any root node $r_i \in R$ to a leaf node $d_i \in D(\mathcal{G})$ equals to the predefined maximal length $L$.

Figure 2 shows examples for generating optimal sequences from two task graphs. Initially, we start with an empty graph. Following the orange example (left), we sample the atomic operation initially, drawing task $t_2$, which induces a new node in $\mathcal{G}^{(orange)}$. When drawing the *and* operation, we first uniformly sample $k = 2$ tasks $\{t_1, t_3\}$. After computing all permutations of the sampled tasks, we get the paths $v_{t_1} \to v_{t_3}$ and $v_{t_3} \to v_{t_1}$ that are appended to the graph structure in steps $\ell = 2$ and $\ell = L = 3$. A practical example of valid task sequences might be grabbing sushi. It is a personal preference whether one first adds wasabi and then soya $(t_2, t_1, t_3)$ or the other way around. Following the green example, we draw the *or* operation and sample tasks $\{t_2, t_3\}$. Subsequently, the atomic operation samples tasks $t_1$ resulting in $\{(t_2, t_1), (t_3, t_1)\}$ at step $\ell = 2$. The generation terminates by appending a singular task, $t_1$ again, and the green trajectory set results in $S^{*(green)} = \{(t_3, t_1, t_1), (t_2, t_1, t_1)\}$. A practical example might be a learning task, where one first improves on Rust or C++ before learning Python and deepens one's understanding of Python.

### 3.2 Universality of the DAG Prior

A directed acyclic graph (DAG) is universal for any finite sequencing problem $S^*$.

**Theorem 1.** *Let $\mathcal{T}$ be a finite set of tasks and $L \in \mathbb{N}$. For any set $S^* \subseteq \mathcal{T}^L$, there exists a directed acyclic graph $\mathcal{G} = (\mathcal{V}, \mathcal{A})$ with $R(\mathcal{G})$ denoting the root nodes, and a mapping function $\phi : \mathcal{V} \to \mathcal{T}$ such that the set of all paths in $\mathcal{G}$ from the root nodes to the leaf nodes are exactly $S^*$ under $\phi$.*

*Proof (Proof By Construction).* Let $S^*$ denote a set of any optimal sequences. For each positional index $i = 1, \ldots, L$ and each task $t \in \mathcal{T}$ that occurs at a position $i$ in at least one sequence $\tau \in S^*$, we create a vertex $v_{t,i}$, i.e., $\mathcal{V} = \{v_{t,i} | \exists \tau \in S^* \ \exists t \in \mathcal{T} : \tau_i = t \land 1 \leq i \leq L\}$. Furthermore, let the mapping function $\phi$ of Definition 1 be $\phi(v_{t,i}) = t$, where a task $t$ might be mapped to several positions within the sequence $\tau$. For every sequence $\tau \in S^*$ and for each positional index $i = 1, \ldots, L-1$, we add edges, i.e., $\mathcal{A} = \{(v_{\tau_i, i}, v_{\tau_{i+1}, i+1}) | \forall \tau \in S^*, i \leq 1 \leq L-1\}$. Hence, we add transitions from one level $i$ to the next $i+1$ preventing any directed cycles. By construction, any node ordering starting from the root nodes $R(\mathcal{G})$ with $\delta^-(v_{t,1}) = \emptyset$ for a node $v_{t,1}$, and traversing directed edges to the leaf nodes, yields a node sequence $(v_{t_1,1}, \ldots v_{t_L,L})$ of length L. Applying, the mapping function $\phi(\cdot)$, we get the task sequence $(\phi(v_{t_1,1}), \ldots \phi(v_{t_L,L}))$ that refers to a $\tau \in S^*$. Hence, it yields valid task paths defined in Definition 2. It follows, we can only traverse edges referring to task sequences that lie in $S^*$ concluding that a DAG is a universal encoder for any finite sequence problem.

Notably, we add edges one-by-one in the proof, a procedure defined by the atomic operator defined in Section 3.1. Due to the monotonicity of expressive power, the theorem trivially still holds for any superset of operators defined in Section 3.1.

### 3.3 Defining the Utility Function $U$

Given a task sequencing problem as $\mathcal{P} := \langle S^*, U \rangle$, we define our utility function as a composition of two distance functions, Dynamic Time Warping (DTW) [25] (cf. Equation (4)) and the Hamming distance [10]. Given two sequences $\tau = (t_1, \ldots, t_L)$ and $\tau' = (t'_1, \ldots, t'_L)$, the Hamming distance $D^H(\tau, \tau')$ between a pair of sequences is the number of positions at which corresponding sequence elements differ as $D^H(\tau, \tau') = \sum_{i=1}^{L} C(t_i, t'_i)$.

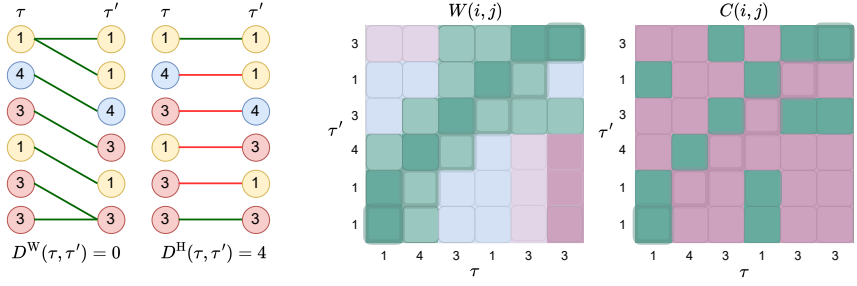

Fig. 3: DTW is able to correctly warp sequences which have shifts of the starting position of the pattern of tasks (e.g. $1, 4, 3, 1, 3$ occur in both $\tau$ and $\tau'$). In contrast, the Hamming distance expects patterns to occur at exact indices on both sequences.

In contrast, DTW computes the warping distance (see Figure 3) between them as $D^{\text{W}}(\tau, \tau') = W(L, L)$ using a recursive function $W$ as:

$$W(i,j) = C(t_i, t'_j) + \min\left(W(i-1, j-1), W(i-1, j), W(i, j-1)\right)$$
$$C(\tau_i, \tau'_j) = \begin{cases} 0 & t_i = t'_j \\ 1 & t_i \neq t'_j \end{cases} \tag{4}$$

We derive a similarity score between two sequences $\tau$ and $\tau'$ to be a function of the two distance functions as shown in Equation 5. We set $\alpha = 0.5$ to have an equal impact from both functions.

$$\text{sim}(\tau, \tau') = \alpha \cdot \frac{1}{1 + D^{\text{W}}(\tau, \tau')} + (1 - \alpha) \cdot \frac{1}{1 + D^{\text{H}}(\tau, \tau')} \tag{5}$$

We then define in Equation 6 the utility function $U(\tau)$ of a trajectory $\tau$ as the highest similarity score with respect to the set of optimal solutions $S^*$ (see Section 3.1). The rationale is that a sequence should have a high utility if it is similar to any of the optimal sequences in $S^*$. For $k \in \{1, \ldots, L\}$, let $\tau_{1:k} := (t_1, \ldots, t_k)$ denote the length-$k$ prefix of $\tau$. We define the prefix utility as

$$U_k(\tau) = \max_{\tau^* \in S^*} \text{sim}(\tau_{1:k}, \tau^*_{1:k}), \quad k \in \{1, \ldots, L\}. \tag{6}$$

This yields the utility vector $U(\tau) := (U_1(\tau), \ldots, U_L(\tau)) \in \mathbb{R}^L_+$.

## 4  Evaluation

In this section, we evaluate our PFTSN method. We assess the effectiveness of our meta-learned task sequencing framework (Section 4.1), compare training paradigms including meta-learning with iterative fine-tuning (Section 4.2), present an ablation study (Section 4.3), and discuss computational costs (Section 4.4). Our model was implemented using PyTorch and was trained on a single Nvidia H100. We open-source our implementation [3] to foster future research.

**Datasets.** Following the data generation step outlined in Section 3.1, we synthesize two benchmark datasets to account for different problem sizes. The components defining a dataset instance are the *number of tasks* $|\mathcal{T}|$, *the maximum number of nodes to add per timestep* $k_{max}$ (which we set to 2) and the *sequence length L*. The *small* dataset ($L = 8$, $|\mathcal{T}| = 8$) and the *large* dataset ($L = 16$, $|\mathcal{T}| = 8$) each consists of 1.000.000 task sequencing problems $\langle S^*, U \rangle$, each containing the optimal trajectories $S^*$, 16 random trajectories $S_{\text{rand}}$ with their utility score $U(S_{\text{rand}})$ and 16 mutated trajectories $S_{\text{mut}}$ with their utility score $U(S_{\text{mut}})$. We further provide separate test and validation sets of 128 unique TS problems for each dataset, each with 16 random trajectories (as initial context) and their utility scores. We utilized multiprocessing across 25 cores on Intel Xeon Platinum 8360Y processors for dataset generation: 2 cores/2 hours for the small dataset and 8 cores/20 hours for the large dataset.

**Baselines.** We compare our approach against a diverse set of task sequencing baselines, including random sampling, a rule-based heuristic, reinforcement learning methods, and large language models (LLMs). We compare the baselines on the test set of TS problems following an iterative sequencing optimization approach, where each method uses the context of sequences so far to predict the optimal sequence (Further information on the evaluation protocol is outlined in Appendix B). Concretely, our method and the baselines are set up as follows:

- **PFTSN.** We train PFTSN separately for every dataset using our method described in Section 2 We summarize additional implementation details and hyper-parameter settings in Appendix C.
- **Random.** Uniformly samples $L$ tasks from $\mathcal{T}$ to form a proposed sequence.
- **Rule-based.** Maintains a task set $\mathcal{T}_i$ for each step $i \in [L]$; it locks in optimal subpaths and removes sampled task IDs from $\mathcal{T}_i$ that do not contribute to an optimal path.
- **Reinforcement learning.** We model trajectory construction as a sequential decision process in which, at each position $t$, the agent selects the next task ID (action) given the current prefix (state). We prefill a replay buffer with the $C \times L$ transitions extracted from the $C$ context trajectories and assign a reward based on the cumulative similarity score. We then train a value-based agent with a periodically synchronized target network and continue training online by generating new trajectories, computing the reward, and adding the resulting transitions back to the buffer. We compare **DDQN** (uniform replay, $\epsilon$-greedy exploration) [28] and **Rainbow** (prioritized replay, $n$-step returns, dueling network, noisy exploration) [11].
- **Large Language Models.** We evaluate state-of-the-art foundation models for inferring optimal task sequences from few-shot trajectory context. Given sampled sequences with similarity-based rewards, the model iteratively proposes an improved length-$L$ sequence. We assess both a base model, **ChatGPT-4.1** [1], and a reasoning model, **DeepSeek-R1-0528-Qwen3-8B** [8]. Implementation details are provided in Appendix D.

---

[3] https://anonymous.4open.science/r/PFTSN-C7B6/README.md

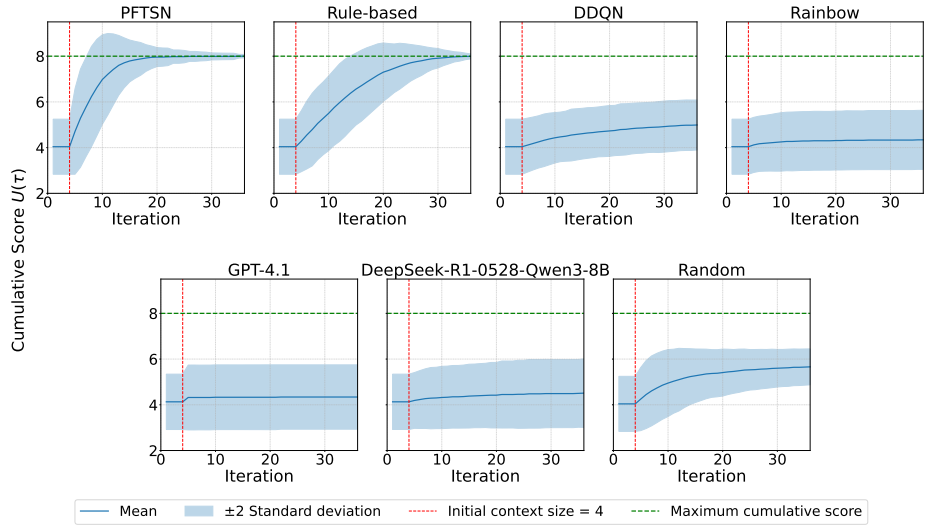

Fig. 4: Cumulative utility scores across all baselines on the small test set sequencing problems after 32 iterations.

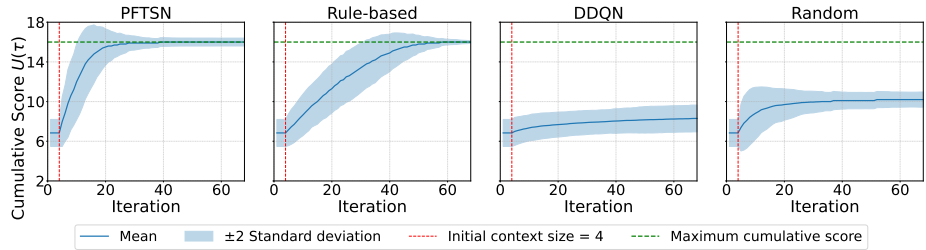

Fig. 5: Cumulative utility scores across all baselines on the large test set sequencing problems after 64 iterations.

### 4.1 Hypothesis 1: Our method outperforms task sequencing baselines that are not meta-learned.

We evaluate all methods on the validation TS problems of the small and large task sequencing benchmarks. For each problem, we initialize every method with an identical random context of four sequences. Methods are allowed 32 iterations on the small dataset and 64 on the large dataset to propose sequences.

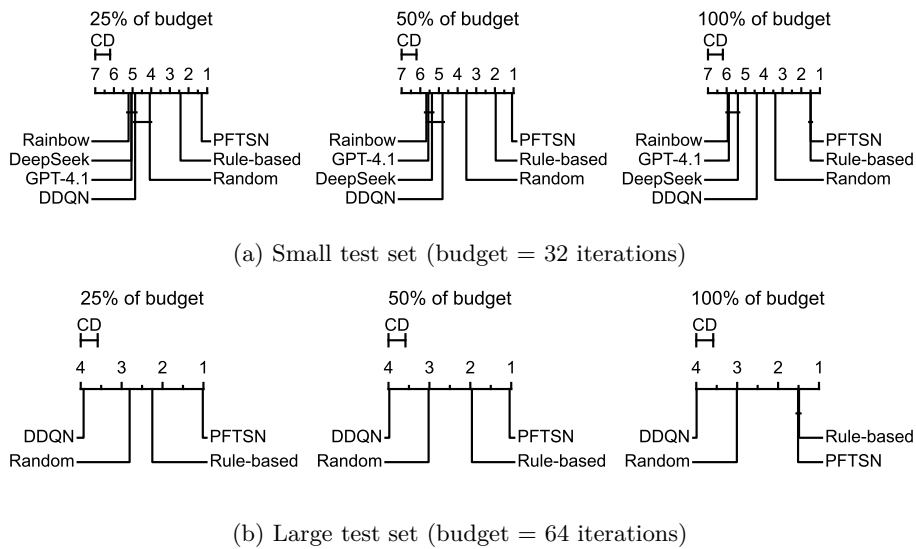

(a) Small test set (budget = 32 iterations)

(b) Large test set (budget = 64 iterations)

Fig. 6: Critical difference diagrams showing the ranks on (a) small and (b) large datasets at 25%, 50%, and 100% of allowed iterations.

Figures 4 show the best-performing trajectories for each method, measured by the cumulative utility $U(\tau)$. Across all settings, PFTSN converges substantially faster than the baselines. PFTSN reaches near-optimal performance, approaching the maximum score of 8 within roughly 20 iterations, whereas the rule-based approach requires around 32 iterations. Random search outperforms the RL baselines, likely because training for only 32 iterations is insufficient to enable adequate exploration and to reliably discover optimal sequences. The LLM baselines show limited improvement: ChatGPT quickly plateaus and repeatedly outputs sequences it has already evaluated, while DeepSeek exhibits only marginal gains over time.

Although the rule-based method achieves lower variance in later iterations, PFTSN consistently demonstrates superior early performance. Figure 5 reports results on the large dataset for the best-performing methods (PFTSN, rule-based, DDQN, and random search), showing a similar pattern. Finally, Figure 6 presents critical difference diagrams for (a) the small dataset and (b) the large dataset, highlighting that PFTSN achieves the highest overall rank. Overall, these results confirm PFTSNs effectiveness in rapidly identifying high-utility task sequences.

### 4.2 Hypothesis 2: Meta-learning is essential for task sequencing problems with limited initial observations.

To further evaluate the necessity of meta-learning for task sequencing, we compare our model in a more extensive comparison by evaluating meta-learning (meta) against iterative fine-tuning (ft) and against no meta-learning. Since the task sequencing problems observed during meta-learning are distinct from the validation problems, we examine if the models can perform better, specifically by training on the current problem's sequences. We therefore evaluate the following baselines: PFTSN (+meta, -ft), PFTSN (+meta, +ft), TSN (-meta, -ft) and TSN (-meta, -ft, +fewer parameters) where TSN refers to our architecture without meta-learning on prior data.

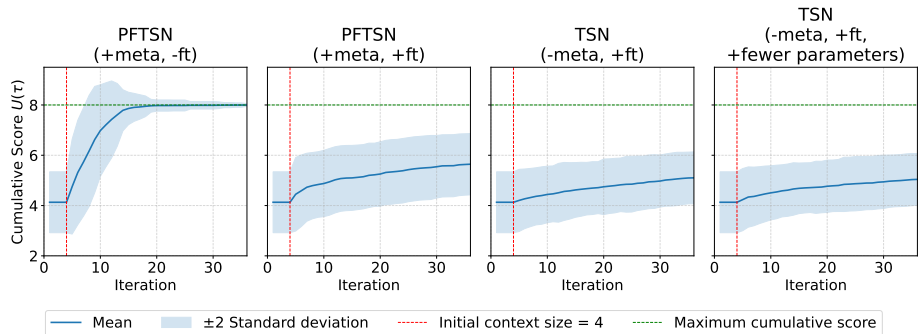

Fig. 7: Training paradigm comparison: Cumulative utility score for 32 iterations on the small dataset.

To implement the fine-tuning baselines, we apply a training protocol designed to handle the limited training data available for the task-sequencing problem, which is constrained by the current context size. We begin by sorting the context based on the sequence's performance (using a cumulative similarity score) and select the top-$k$ sequences as our targets (for all experiments, we use $k = 5$). For each target sequence, we identify all lower-performing sequences in the context. We then create multiple training examples by considering every possible non-empty combination of these lower-performing sequences. After training, we iteratively generate and evaluate new sequences, adding each to our growing context. As the number of possible training examples grows exponentially, we switch to a sampling-based training data generation approach once the possible number of potential training examples exceeds our predefined batch size. This process continues until all required sequences are generated and evaluated.

We evaluate our method against the baselines on the 128 task sequencing problems on the small benchmark for 32 iterations. Our meta-learned method outperforms all fine-tuned baselines (cf. Figure 7) in terms of convergence. We find that fine-tuning our meta-learned model on sub-optimal sequences hinders rather than improves performance, suggesting that the meta-learned representations are already well-suited for identifying optimal sequences.

### 4.3  Ablation Study

We conduct two targeted ablations to assess robustness. First, we vary the amount of initial context available at evaluation time (Figure 8) and observe almost no performance degradation as the context is reduced, indicating that our method remains effective even with very limited initial information. Second, we examine the effect of the maximum context length used during meta-learning (8 vs. 16; Figure 1, Appendix E) and find performance to be nearly identical, suggesting that our approach does not depend on long context windows during training.

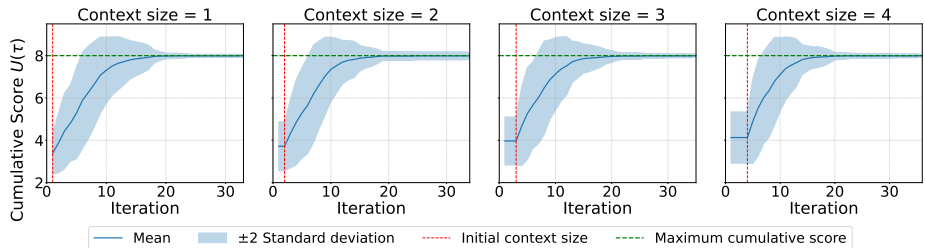

Fig. 8: Ablating the initial context available during evaluation for our proposed method. There is no decrease in performance as we decrease the available initial context. The plots of evaluation performance show that our method is robust to changing amounts of information in the initial context.

### 4.4 Computational Costs

We further evaluate the computational costs of our proposed method against the baselines, as shown in Figure 9. The evaluation was conducted on a single NVIDIA A40 graphics card[4], performing 32 iterations for 8 task-sequencing problems on the small benchmark. While our method requires a larger computational time than the baselines, it delivers substantially superior performance as demonstrated in Figures 4, 7 and 8. This experiment measures only the runtime of generating the next sequence, assuming task execution has no cost. In real-world applications, task execution introduces additional overhead, making the relative runtime increase of our method less significant in practice (a more detailed analysis is provided in Appendix G). Given the significant quality improvements achieved, we believe this computational overhead represents a favorable performance-accuracy trade-off.

## 5    Related Work

While initial work on task sequencing and curriculum learning focused on heuristic rules such as training on easier examples first [26,2] or harder examples first [3], subsequent research developed curricula capable of detecting example difficulty [34,9]. In reinforcement learning, curriculum methods often recommend the next task online [15,23]. Continual learning work also highlights task sequencing as key for mitigating catastrophic forgetting [20,21]. In contrast, our method removes the requirement of expert-designed curricula and sequences tasks in a domain-agnostic manner.

---
[4] Except for the LLM baselines; further details are provided in Appendix F

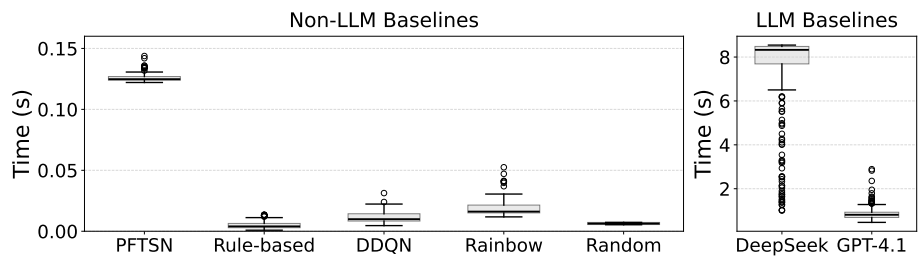

Fig. 9: Comparison of PFTSN and baseline inference runtimes for sequence generation. Results are separated into non-LLM and LLM models, as LLM inference times are on a different order of magnitude (in seconds).

Meta-learning approaches for curriculum teaching and task sequencing have demonstrated superior performance compared to handcrafted and online sequencing methods. CMDP [19] employs a meta-learned DQN teacher that leverages both environment-specific features and the student's policy parameters to optimize learning sequences. Further, MMACL [31] metalearns a curriculum teacher that utilizes task descriptors such as the start and goal distances and, from them, predicts which navigation task should be trained next. AGAIN [24] meta-learns a curriculum techer pre-train with ALP-GMM [23], find the closest prior student via k-NN, then mix that students GMM curriculum with ongoing ALP-GMM sampling. In addition, meta-learning has been explored for student-teacher paradigms [6]. Despite their effectiveness, these approaches share a common limitation: they depend on environment-specific task embeddings and exploit particular structures within these embeddings. A related approach in terms of learning from generated priors is PFN [17], however, that work focuses on the orthogonal problem of plain supervised learning. None of the prior work demonstrated the ability of TS methods to meta-learn from a large set of TS problems and generalize to novel sequencing problems in a few-shot modality.

## 6  Conclusion

While task sequence is an important problem for the Machine Learning community, no prior work has studied the capabilities of deep neural networks in meta-learning from a large set of TS problems. In this paper, we propose a novel framework that unifies meta-learning benefits for task sequencing by training a transformer model to predict optimal curricula for a given context. Whereas traditional TS approaches rely on domain-specific information and lack adaptability to a broad range of task distributions, our model generalizes to future TS problems in a few-shot modality. Experimental results in a controlled environment demonstrate that our model generates context-sensitive optimal sequences and outperforms baselines like random, rule-based, RL and large language model baselines.

## 7   Limitations

In this study, we investigated a foundation model for sequencing learned on synthesized priors. Having established an important proof-of-concept and demonstrated the feasibility of meta-learning for few-shot task sequencing, we will extend our work to real-world priors in the future. Furthermore, we expect our approach to be particularly suitable for practical settings in which task execution introduces substantial overhead compared to recommendation time.

**Acknowledgments.**

## Acknowledgments

Josif Grabocka and Arlind Kadra acknowledge the funding support from the Bayerisches Landesamt fur Steuer for the Bavarian AI Taxation Laboratory.

Gresa Shala and Josif Grabocka acknowledge the funding by The Carl Zeiss Foundation through the research network Responsive and Scalable Learning for Robots Assisting Humans (ReScaLe) of the University of Freiburg.

Josif Grabocka acknowledges the financial support from the Albanian-American Development Foundation through the READ program.

Moreover, we gratefully acknowledge the scientific support and HPC resources provided by the Erlangen National High Performance Computing Center (NHR@FAU) of the Friedrich-Alexander-Universität Erlangen-Nürnberg (FAU). NHR@FAU hardware is partially funded by the German Research Foundation (DFG)  440719683.

## References

1. Achiam, J., Adler, S., Agarwal, S., Ahmad, L., Akkaya, I., Aleman, F.L., Almeida, D., Altenschmidt, J., Altman, S., Anadkat, S., et al.: Gpt-4 technical report. arXiv preprint arXiv:2303.08774 (2023)
2. Bengio, Y., Louradour, J., Collobert, R., Weston, J.: Curriculum learning. In: Proc. of the Int. Conference on Machine Learning (ICML) (2009)
3. Cho, M., Park, J., Lee, S., Sung, Y.: Hard tasks first: Multi-task reinforcement learning through task scheduling. In: Proc. of the Int. Conference on Machine Learning (ICML) (2024)
4. Defazio, A., Yang, X., Khaled, A., Mishchenko, K., Mehta, H., Cutkosky, A.: The road less scheduled. In: Advances in Neural Information Processing Systems (NeurIPS) (2024)
5. Ericsson, K.A., Krampe, R.T., Tesch-Römer, C.: The role of deliberate practice in the acquisition of expert performance. Psychological Review **100**(3), 363–406 (1993)
6. Fan, Y., Tian, F., Qin, T., Li, X.Y., Liu, T.Y.: Learning to teach. In: Proc. of the Int. Conf. on Learning Representations (ICLR) (2018)

7. Ge, Y., Zhu, F., Chen, D., Zhao, R., Li, H.: Self-paced contrastive learning with hybrid memory for domain adaptive object re-id. In: Advances in Neural Information Processing Systems (NeurIPS) (2020)
8. Guo, D., Yang, D., Zhang, H., Song, J., Wang, P., Zhu, Q., Xu, R., Zhang, R., Ma, S., Bi, X., et al.: Deepseek-r1: Incentivizing reasoning capability in llms via reinforcement learning. arXiv preprint arXiv:2501.12948 (2025)
9. Guo, S., Huang, W., Zhang, H., Zhuang, C., Dong, D., Scott, M.R., Huang, D.: Curriculumnet: Weakly supervised learning from large-scale web images. In: Proc. of European Conference on Computer Vision (ECCV) (2018)
10. Hamming, R.W.: Error detecting and error correcting codes. In: The Bell System Technical Journal (1950)
11. Hessel, M., Modayil, J., Van Hasselt, H., Schaul, T., Ostrovski, G., Dabney, W., Horgan, D., Piot, B., Azar, M., Silver, D.: Rainbow: Combining improvements in deep reinforcement learning. In: Proceedings of the AAAI conference on artificial intelligence. vol. 32 (2018)
12. Hunicke, R., LeBlanc, M., Zubek, R.: Mda: A formal approach to game design and game research. In: Proceedings of the AAAI Workshop on Challenges in Game AI (2004)
13. Kravchenko, A., Cusack, R.: The limitations of automatically generated curricula for continual learning. In: PLOS ONE (2024)
14. Loshchilov, I., Hutter, F.: Decoupled weight decay regularization. In: Proc. of the Int. Conf. on Learning Representations (ICLR) (2017)
15. Matiisen, T., Oliver, A., Cohen, T., Schulman, J.: Teacher–student curriculum learning. In: IEEE transactions on neural networks and learning systems (2019)
16. Mukherjee, S., Mitra, A., Jawahar, G., Agarwal, S., Palangi, H., Awadallah, A.: Orca: Progressive learning from complex explanation traces of gpt-4 (2023), *arXiv:2306.02707*
17. Müller, S., Hollmann, N., Arango, S.P., Grabocka, J., Hutter, F.: Transformers can do bayesian inference. In: Proc. of the Int. Conf. on Learning Representations (ICLR) (2022)
18. Narvekar, S., Peng, B., Leonetti, M., Sinapov, J., Taylor, M.E., Stone, P.: Curriculum learning for reinforcement learning domains: a framework and survey. In: Journal of Machine Learning Research (JMLR) (2020)
19. Narvekar, S., Stone, P.: Generalizing curricula for reinforcement learning. In: Proc. of the Int. Conference on Machine Learning (ICML) (2020)
20. Nguyen, C.V., Achille, A., Lam, M., Hassner, T., Mahadevan, V., Soatto, S.: Toward understanding catastrophic forgetting in continual learning (2019), *arXiv:1908.01091* [cs.LG]
21. Nguyen, T., Nguyen, C., Pham, Q., Nguyen, B., Ramasamy, S., Li, X., Nguyen, C.: Sequence transferability and task order selection in continual learning (2025), *arXiv:2502.06544*
22. Platanios, E.A., Stretcu, O., Neubig, G., Poczos, B., Mitchell, T.: Competence-based curriculum learning for neural machine translation. In: Proceedings of the 2019 Conference of the North American Chapter of the Association for Computational Linguistics (2019)
23. Portelas, R., Colas, C., Hofmann, K., Oudeyer, P.Y.: Teacher algorithms for curriculum learning of deep rl in continuously parameterized environments. In: Proc. of the Conf. on Robot Learning (CoRL) (2020)
24. Portelas, R., Romac, C., Hofmann, K., Oudeyer, P.Y.: Meta automatic curriculum learning (2020), *arXiv:2011.08463*

25. Sakoe, H., Chiba, S.: Dynamic programming algorithm optimization for spoken word recognition. In: IEEE Transactions on Audio, Speech, and Language Processing (1978)
26. Sanger, T.D.: Neural network learning control of robot manipulators using gradually increasing task difficulty. In: IEEE Robotics & Automation Magazine (1994)
27. Shazeer, N.: Glu variants improve transformer (2020), *arXiv:2002.05202* [cs.LG]
28. Van Hasselt, H., Guez, A., Silver, D.: Deep reinforcement learning with double q-learning. In: Proc. of the National Conference on Artificial Intelligence (AAAI) (2016)
29. Vaswani, A., Shazeer, N., Parmar, N., Uszkoreit, J., Jones, L., Gomez, A.N., Kaiser, Ł., Polosukhin, I.: Attention is all you need. In: Advances in Neural Information Processing Systems (NeurIPS) (2017)
30. Wilson, C.E.: Taking usability practitioners to task. In: Interactions (2007)
31. Xu, Z., Zhang, Y., Shperberg, S.S., Mirsky, R., Jiang, Y., Liu, B., Stone, P.: Model-based meta automatic curriculum learning. In: Conference on Lifelong Learning Agents (CoLLAs) (2023)
32. Yin, Y., Chen, Z., Liu, G., Yin, J., Guo, J.: Autonomous navigation of mobile robots in unknown environments using off-policy reinforcement learning with curriculum learning. Expert Systems with Applications (2024)
33. Zhang, B., Sennrich, R.: Root mean square layer normalization. In: Advances in Neural Information Processing Systems (NeurIPS) (2019)
34. Zhang, D., Meng, D., Han, J.: Co-saliency detection via a self-paced multiple-instance learning framework. In: IEEE Transactions on Pattern Analysis and Machine Intelligence (2017)

# Supplementary materials for : Learning to Order: Task Sequencing as In-Context Optimization

This appendix provides additional details supporting the methodology and experiments presented in the main paper Learning to Order: Task Sequencing as In-Context Optimization. We describe the generation of mutated trajectories used for training (in Appendix A), the evaluation protocol used to benchmark all methods (in Appendix B), and implementation details including hyperparameter settings (in Appendix C) and LLM prompting setup (in Appendix D). We further present ablation studies analyzing key design choices (in Appendix E) as well as details on the timing measurements and runtime trade-offs of the LLM baselines (in Appendices F and G). Finally, the use of LLMs in preparing the main paper is outlined (in Appendix H).

## A  Generating Mutate($S_{\text{mut}}$)

We generate mutated trajectories Mutate($S_{\text{mut}}$) by sampling a random optimal sequence $\tau^* \sim S^*$. We then sample a resampling integer $L_{\text{mut}} \sim \text{Uniform}([L/2], L)$. The first $L_{\text{mut}}$ elements of $\tau^*$ are preserved, while the remaining elements are replaced with randomly drawn tasks, generating trajectories with partial preservation of $S^*$.

## B  Evaluation Protocol

In a series of iterations, each technique uses the context of observed sequences and their utility series values to predict the next optimal sequence. The predicted sequence and its utility values are appended to the context, and the optimization moves to the next iteration. We run each method on each test TS problem for 32 iterations, given 4 initial random sequences (i.e., each method recommends 32 sequences step by step). In the end, we select the best sequence discovered by each baseline, where the utility of the sequence is the sum of utility values for each partial sub-sequence of tasks (from 1 to $L$).

## C  Implementation Details and Hyperparameter Settings

During training, the generated sequences are not appended to the current context; instead, we sample a new time series (TS) problem $\langle S^*, U \rangle$ for each iteration. In contrast, during evaluation, after generating and evaluating a new sequence, it is appended to the current context. To ensure that our model performs effectively with varying context sizes $C$, we sample a random context size $C \sim \text{Uniform}(C_{\min}, C_{\max})$ for each training iteration. To accommodate different quality levels of context, we implement strategic data mixing. As previously described, our data comprises two sources: random sequences $S_{\text{rand}}$ and mutated sequences $S_{\text{mut}}$. During the initial evaluation phase, we begin with a relatively small random context (due to limited knowledge about the current TS problem). However, as the evaluation progresses, the context expands as our model iteratively generates increasingly optimal sequences. We therefore dynamically determine the proportion of mutated trajectories using $C_{\text{mut}} = \left\lfloor C \cdot \frac{C - C_{\min}}{C_{\max} - C_{\min}} \right\rfloor$ and random trajectories using $C_{\text{rand}} = C - C_{\text{mut}}$. We then sample $C_{\text{mut}}$ sequences from $S_{\text{mut}}$ and $C_{\text{rand}}$ sequences from $S_{\text{rand}}$. This adaptive context mixing approach ensures a proportional increase in samples drawn from $S_{\text{mut}}$ as $C$ approaches $C_{\max}$.

Building on this adaptive context mixing strategy and the model architecture outlined in Section 2, we train our model from scratch on each dataset using the hyper-parameters detailed in Table 1. For reference, it required 2 hours (small) and 8 hours (large) to train our model on an H100 with this setup.

Table 1: Hyperparameters for the Prior-Fitted TS Networks.

| Hyperparameter | Value |
| --- | --- |
| Training Iterations | 20,000 |
| Batch Size | 1,024 |
| Learning Rate | $1.0 \times 10^{-3}$ |
| Learning Rate Scheduler | Schedule-Free [4] |
| Weight Decay | 1e-3 |
| Learning Rate Warmup | 500 |
| Optimizer | AdamW [14] |
| $C_{min}, C_{max}$ | 4, 16 |
| Number of Transformer Blocks | 12 |
| Number of Attention Heads | 8 |
| Dropout Rate | 0.05 |
| Embedding Size ($d_{emb}$) | 64 |
| Hidden Layer Size | 256 |
| Temperature Scaling | 4 |

## D  Implementation Details LLM

For GPT-4.1, we used the OpenAI API. DeepSeek was self-hosted locally using the SGLang framework on four NVIDIA H100 GPUs. At each iteration, the LLM is provided with the prompt shown below. The placeholders `{L}` (sequence length) and `{context}` are replaced at call time, while the remainder of the prompt is

*Prompt:*

```
## ROLE
You are SequenceOptimizer, designed to generate optimal task
sequences through few-shot learning. Given suboptimal context
task sequences and their performance metrics (similarity scores),
analyze these examples and propose an improved sequence that
maximizes overall performance.

## INPUT
You receive a context tensor of up to N previous trials (rows),
each of fixed sequence length L = :

context_tensor ∈ ℝ^[N x L x 2]
    [:,:,0]   task-ID at timestep j in trial n   (integer, 0-7)
    [:,:,1]   similarity score at timestep j     (float,   0-1)

The similarity score at timestep j represents the best match
between the current partial sequence [0,...,j] and any prefix
of the unknown target sequences (max over all targets).

## OBJECTIVE
Inspect the (Task-ID, similarity score) histories, infer which
subsequences or positional patterns yield higher scores, and emit
one new length-L sequence of exactly {L} integers (each in 0-7)
that achieves a similarity score of 1.0 at every timestep.

## CONSTRAINTS
  * Sequence length  L = {L}
  * Each element ∈  {0, 1, 2, 3, 4, 5, 6, 7}
  * Return exactly   {L} integers, space-separated, no extra tokens

## EXAMPLE  (arbitrary -- focus solely on the context below)
  Target sequence:  [4, 2, 1, 5]
  c_1 = [4,2,2,3]  scores [1,1,0,0]  (first two are perfect matches)
  c_2 = [0,1,2,3]  scores [0,0,0,0]  (lower performance)
  Perfect match would yield scores [1,1,1,1].

## CONTEXT
{context}

## OUTPUT
Return exactly {L} integers, space-separated.
No headers, comments, punctuation, or extra tokens.
```

## E  Ablations

We evaluate the sensitivity of our approach to key design choices through two ablations (see Figure 8 and Figure 1). Figure 1 investigates the importance of our chosen maximum context length by repeating meta-learning with maximum lengths of 8 and 16 steps. As shown, performance with a context length of 8 matches that of 16, demonstrating that our method remains robust even when meta-learned with a much shorter context horizon.

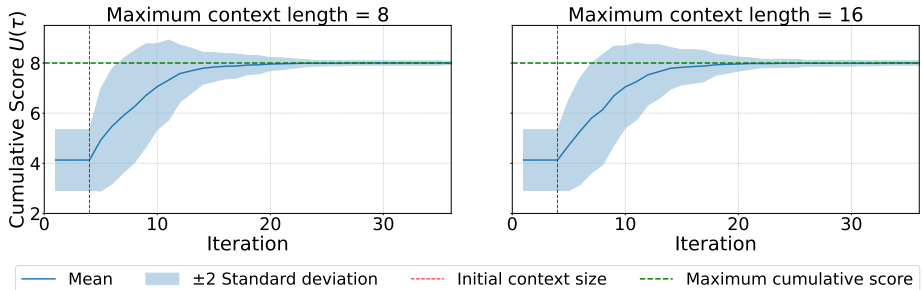

Fig. 1: Ablating the maximum context length during meta-learning for our proposed method. We ablate using a maximum context length of 8 and 16 during meta-learning. The plots of the evaluation performance show that our method is robust to using a shorter maximum context length of 8 during meta-learning.

## F  Implementation details for the timing of the LLM Baselines.

We deployed DeepSeek locally using the SGLang inference framework on a server equipped with four NVIDIA H100 GPUs. Inference requests were issued from a separate local client machine. To ensure that our latency measurements do not include network overhead, we aggregate two timing values: (1) the end-to-end runtime measured on the client machine, which includes prompt preprocessing, and (2) the server-side runtime measured directly on the SGLang server. For the server-side timing, we rely on the logging metrics provided by SGLang and report the e2e latency value for each request, defined as the time from when the server receives the prompt to when output generation is complete. This measurement isolates inference performance by excluding network communication costs.

## G    Analysis of Inference and Execution Costs

There are two components influencing the overall runtime. First, the time required for the sequencing model to recommend the next sequence to evaluate. Second, the time required to execute all tasks within a sequence. While PFTSN is slower in recommending the next sequence, it converges in fewer iterations than the rule based method. On the small dataset, PFTSN converges after 19 iterations compared to 29 for the rule based method. On the large dataset, it converges after 29 iterations compared to 58.

To illustrate the trade off, the inference cost on the small dataset is 0.13 seconds per iteration for PFTSN and 0.007 seconds for the rule based method, resulting in total inference times of 2.47 seconds and 0.203 seconds. Let $x$ denote the time required to execute a sequence. Both methods have equal total runtime when
$$19x + 2.47 = 29x + 0.203,$$
yielding $x = 0.2267$ seconds. Therefore, whenever executing a sequence requires more than 0.2267 seconds, PFTSN achieves a lower overall runtime.

## H    LLM usage

In accordance with ECML 2026's author guidelines on the use of Large Language Models, we disclose that ChatGPT and Claude were employed for light editing purposes, including correcting grammatical mistakes and improving the phrasing of select sentences in the main text. They were also used to assist with straightforward coding tasks, such as generating visualizations. LLMs played no role in the conception of research ideas, the design of experiments, the execution of analyses, or the drafting of substantive content. All research contributions, analyses, and conclusions are solely the original work of the authors.